\documentclass[11pt,a4paper]{article}
\usepackage[hyperref]{naaclhlt2018}
\usepackage{latexsym}

\usepackage{url}

\usepackage{times}
\usepackage{latexsym}
\usepackage{graphicx}
\usepackage{caption}
\usepackage{subcaption}
\usepackage{algorithm}
\usepackage{algpseudocode}
\usepackage{amsmath}
\usepackage{amsfonts}
\usepackage{mdwlist}
\usepackage{url}
\usepackage{tikz}
\usepackage{float}

\aclfinalcopy 

\newcommand{\epool}{$\textrm{entPool}_c$}
\newcommand{\ppool}{$\textrm{patPool}_c$}
\newcommand{\seed}{$\textrm{seeds}_c$}

\title{Lightly-supervised Representation Learning with Global Interpretability}

\author{Marco A. Valenzuela-Esc\'{a}rcega, Ajay Nagesh, and Mihai Surdeanu \\
University of Arizona, Tucson, AZ, USA \\
  {\tt \{marcov, ajaynagesh, msurdeanu\}@email.arizona.edu}}

\date{}

\begin{document}
\maketitle
\begin{abstract}
We propose a lightly-supervised approach for information extraction, in particular named entity classification, which combines the benefits of traditional bootstrapping, i.e., use of limited annotations and interpretability of extraction patterns, with the robust learning approaches proposed in representation learning.
Our algorithm iteratively learns custom embeddings for both the multi-word entities to be extracted and the patterns that match them from a few example entities per category.
We demonstrate that this representation-based  approach outperforms three other state-of-the-art bootstrapping approaches on two datasets: CoNLL-2003 and OntoNotes. 
Additionally, using these embeddings, our approach outputs a globally-interpretable model consisting of a 
decision list, by ranking patterns based on their proximity to the average entity embedding in a given class.
We show that this interpretable model performs close to our complete bootstrapping model, proving  that representation learning can be used to produce interpretable models with small loss in performance. 
\end{abstract}

\section{Introduction}

One strategy for mitigating the cost of supervised learning in information extraction (IE) is to bootstrap extractors
with light supervision from a few provided examples (or seeds).
Traditionally, bootstrapping approaches iterate between learning extraction patterns such as word $n$-grams, e.g., the pattern ``{\tt @ENTITY , former president}'' could be used to extract person names,\footnote{In this work we use surface patterns, but the proposed algorithm is agnostic to the types of patterns learned.} and applying these patterns to extract the desired structures (entities, relations, etc.)~\cite[inter alia]{carlson2010coupled,gupta2014improved,gupta2015distributed}. One advantage
of this direction is that these patterns are interpretable, 
which mitigates the maintenance cost associated with machine learning systems~\cite{43146}.

On the other hand, representation learning has proven to be useful for natural
language processing (NLP) applications~\cite[inter alia]{mikolov2013distributed,riedel2013relation,toutanova2015representing,toutanova2016compositional}. 
Representation learning approaches often include a component that is trained in an unsupervised manner, e.g., predicting words based on their context from large amounts of data, 
mitigating the brittle statistics 
affecting traditional bootstrapping approaches. However, the resulting real-valued embedding vectors are hard to interpret. 

Here we argue that these two directions are complementary, and should be combined.
We propose such a bootstrapping approach for information extraction (IE), which blends the advantages of both directions. As a use case, we instantiate our idea for named entity classification (NEC), i.e., classifying a given set of unknown entities
 into a predefined set of categories~\cite{collins99}. 
The contributions of this work are: 
 
{\flushleft {\bf (1)}} We propose an approach for bootstrapping NEC that iteratively  learns custom embeddings for {\em both} the multi-word entities to be extracted and the patterns that match them from a few example entities per category.
Our approach changes the objective function of a neural network language models (NNLM) to include a semi-supervised component that models the known examples, i.e., by {\em attracting} entities and patterns in the same category to each other and {\em repelling} them from elements in different categories, and it adds an external iterative process that ``cautiously'' augments the pools of known examples~\cite{collins99}. 

{\flushleft {\bf (2)}}  We demonstrate that our representation learning approach is
suitable for semi-supervised NEC. We compare our approach against several state-of-the-art semi-supervised approaches on two datasets: CoNLL-2003~\cite{conll_dataset} and OntoNotes~\cite{ontonotes}. 
We show that, despite its simplicity, our method outperforms all other approaches. 

{\flushleft {\bf (3)}}  
Our approach also outputs an interpretation of the learned model, consisting of a decision list of patterns, where each pattern gets a score per class based on the proximity of its embedding to the average entity embedding in the given class. This interpretation is global, i.e., it explains the entire model rather than local predictions. 
We show that this decision-list model performs comparably to the complete model on the two datasets. This guarantees that the resulting system can be understood, debugged, and maintained by non-machine learning experts. 
 Further, this model outperforms considerably an interpretable model that uses pretrained embeddings, demonstrating that our custom embeddings help interpretability.

\section{Related Work}

Bootstrapping is an iterative process that alternates between learning representative patterns, and acquiring new entities (or relations) belonging to a given category~\cite{riloff1996automatically,mcintosh2010unsupervised}. 
Patterns and extractions are ranked using either formulas that measure their frequency 
and association with a category, 
or classifiers, which increases robustness due to regularization~\cite{carlson2010coupled,gupta2015distributed}.

Distributed representations of words~\cite{mikolov2013distributed,levy2014dependency} serve as underlying representation for many NLP tasks such as information extraction and question answering \cite{riedel2013relation,toutanova2015representing,toutanova2016compositional,Sharp2016}.
However, most of these works that customize embeddings for a specific task rely on some form of supervision. 
In contrast, our approach is lightly supervised, with a only few seed examples per category. 
\citet{batista2015emnlp}  perform bootstrapping for relation extraction using pre-trained word embeddings.  
They do not learn custom pattern embeddings that apply to multi-word entities and patterns. We show that customizing embeddings for the learned patterns is important for interpretability. 

Recent work has focused on explanations of machine learning models that are model-agnostic but local, i.e., they interpret individual model predictions \cite{anchors:aaai18,lime:kdd16}. 
In contrast, our work produces a global interpretation, which explains the entire extraction model rather than individual decisions. 

Lastly, our work addresses the interpretability aspect of information extraction methods.
Interpretable models mitigate
the technical debt of machine learning~\cite{43146}. For example, it allows
domain experts to make manual, gradual improvements to the models. This is why
rule-based approaches are commonly used in industry applications, where software maintenance is
crucial~\cite{chiticariu2013rule}. 
Furthermore, the need for interpretability also arises in critical systems, e.g., recommending treatment to patients, where these systems are deployed to aid human decision makers~\cite{rudin17}.    
The benefits of interpretability have
encouraged efforts to either extract interpretable models from opaque
ones~\cite{craven1996extracting}, or to explain their
decisions~\cite{ribeiro2016should}. 

As machine learning models are becoming more
complex, the focus on interpretability has become more important, with new
funding programs focused on this topic.\footnote{
DARPA's
  Explainable 
  AI program: {\scriptsize
    \url{http://www.darpa.mil/program/explainable-artificial-intelligence}}.}
    Our approach for exporting an interpretable model (\S\ref{sec:interp}) is similar to \citet{valenzuela2016snaptogrid}, but we start from distributed representations, whereas they started from a logistic regression model with explicit features.

\section{Approach}

\subsection*{Bootstrapping with representation learning}
\label{sec:embed}

Our algorithm iteratively grows a pool of multi-word entities (\epool) and $n$-gram patterns (\ppool) for each category of interest $c$, and learns custom embeddings for both, which we will show are crucial for both performance and interpretability. 

The entity pools are initialized with a few seed examples (\seed) for each category.
For example, in our experiments we initialize the pool for a {\small {\tt person names}} category with 10 names such as {\em Mother Teresa}. 
Then the algorithm iteratively applies the following three steps for $T$ epochs:
{\flushleft {\bf (1)} {\em Learning custom embeddings}}: The algorithm learns custom embeddings for all entities and patterns in the dataset, using the current {\epool}s as supervision. This is a key contribution, and is detailed in the second part of this section. 
{\flushleft {\bf (2)} {\em Pattern promotion}}: We generate the patterns that match the entities in each pool \epool, rank those patterns using point-wise mutual information (PMI) with the corresponding category, and select the top ranked patterns for promotion to the corresponding pattern pool \ppool. {\flushleft {\bf (3)} {\em Entity promotion}}: Entities are promoted to \epool\ using a multi-class classifier that estimates the likelihood of an entity belonging to each class~\cite{gupta2015distributed}. 
Our feature set includes, for each category $c$: (a) edit distance between the candidate entity $e$ and current $e_c$s $\in$ \epool, (b) 
the PMI (with $c$) of the patterns in \ppool\ that matched $e$ in the training documents, 
and (c) similarity between $e$ and $e_c$s in a semantic space. 
For the latter feature group, we use two sets of vector representations for entities. 
The first is the set of embedding vectors learned in step (1). The second includes pre-trained word embeddings; for multi-word entities and patterns, we simply average the embeddings of the component words.
We use these vectors to compute the cosine similarity score of a given candidate entity $e$ to the entities in \epool, and add the average and maximum similarities as features.
The top 10 entities classified with the highest confidence for each class are promoted to the corresponding \epool\ after each epoch.

\subsection*{Learning custom embeddings}

We train our embeddings for both entities and patterns by maximizing the objective function $J$:
\begin{equation}
\small
\label{eq:J}
  J = \text{SG} + \text{Attract} + \text{Repel}
\end{equation}
where SG, Attract, and Repel are individual components of the objective function designed to model both the unsupervised, language model part of the task as well as the light supervision coming from the seed examples, as detailed below.

The SG term captures the original objective function of the {\em S}kip-{\em G}ram model of ~\citet{mikolov2013distributed}, but, crucially, adapted to operate over multi-word entities and contexts consisting not of bags of context words, but of the patterns that match each entity:
\begin{equation}
\small
  \label{eq:rank}
    \begin{aligned}
	\text{SG} = \sum_{e} [ \log(\sigma(V_e^\top V_{pp})) + \sum_{np} \log(\sigma(-V_e^\top V_{np})) ]
    \end{aligned}
\end{equation}
where $e$ represents an entity, $pp$ represents a positive pattern, i.e., a pattern that matches entity $e$ in the training texts, $np$
represents a negative pattern, i.e., it has not been seen with this entity,
and $\sigma$ is the sigmoid function.
Intuitively, this component forces the embeddings of entities to be similar to the embeddings of the patterns that match them, and dissimilar to the negative patterns. 

The second component, Attract, encourages entities or patterns in the same pool to be close to each other. For
example, if we have two entities in the pool known to be person names, they
should be close to each other in the embedding space:
\begin{equation}
\small
  \text{Attract} = \sum_{P} \sum_{x1,x2 \in P} \log(\sigma(V_{x1}^\top V_{x2}))
\end{equation}
where $P$ is the entity/pattern pool for a category, and $x1,x2$ are entities/patterns in said pool.

Lastly, the third term, Repel, encourages that the pools be mutually exclusive, which is a soft version of the counter training approach of~\citet{yangarber03} or the weighted mutual-exclusive bootstrapping algorithm of~\citet{mcintosh2008weighted}. For example,
person names should be far from organization names in the semantic embedding space:
\begin{equation}
\small
  \text{Repel} = \sum_{P1,P2 \text{ if } P1 \ne P2} \sum_{x1 \in P1} \sum_{x2 \in P2} \log(\sigma(-V_{x1}^\top V_{x2}))
\end{equation}
where $P1,P2$ are different pools, and $x1$ and $x2$ are entities/patterns in $P1$, and $P2$, respectively.

We term the complete algorithm that learns and uses custom embeddings as {\em Emboot} ({\em Em}beddings for {\em boot}strapping) and the stripped-down version without them as {\em EPB} ({\em E}xplicit {\em P}attern-based {\em B}ootstrapping).\footnote{EPB is similar to \cite{gupta2015distributed}; the main difference is that we use pretrained embeddings in the entity promotion classifier rather than Brown clusters.}

\subsection*{Interpretable model}

\label{sec:interp}

In addition of its output (\epool{s}), Emboot produces custom entity and pattern embeddings that can be used to construct a a decision-list model, which provides a global, deterministic interpretation of what Emboot learned. 

This interpretable model is constructed as follows.
First, we produce an average embedding per category by averaging the embeddings of the entities in each \epool. 
Second, we estimate the cosine similarity between each of the pattern embeddings and these category embeddings, and convert them to a probability distribution using a softmax function; $prob_c(p)$ is the resulting probability of pattern $p$ for class $c$.
Third, each candidate entity to be classified, $e$, receives a score for a given class $c$ from all patterns in \ppool\ that match it.
The entity score aggregates the relevant pattern probabilities using Noisy-Or:

\begin{scriptsize}
\begin{equation}
\small
    Score(e, c) = 1 - \prod_{\{ p_c \in patPool_c \mid matches(p_c, e) \}} (1 - prob_c(p_c))
\end{equation}
\end{scriptsize}Each entity is then assigned to the category with the highest overall score.

\section{Experiments}
\label{sec:experiments}

\begin{figure*}[t!]
  \centering
  \vspace*{-4mm}
  \subcaptionbox{Embeddings initialized randomly\label{fig:bs_epoch_0}}{\includegraphics[width=0.3\textwidth]{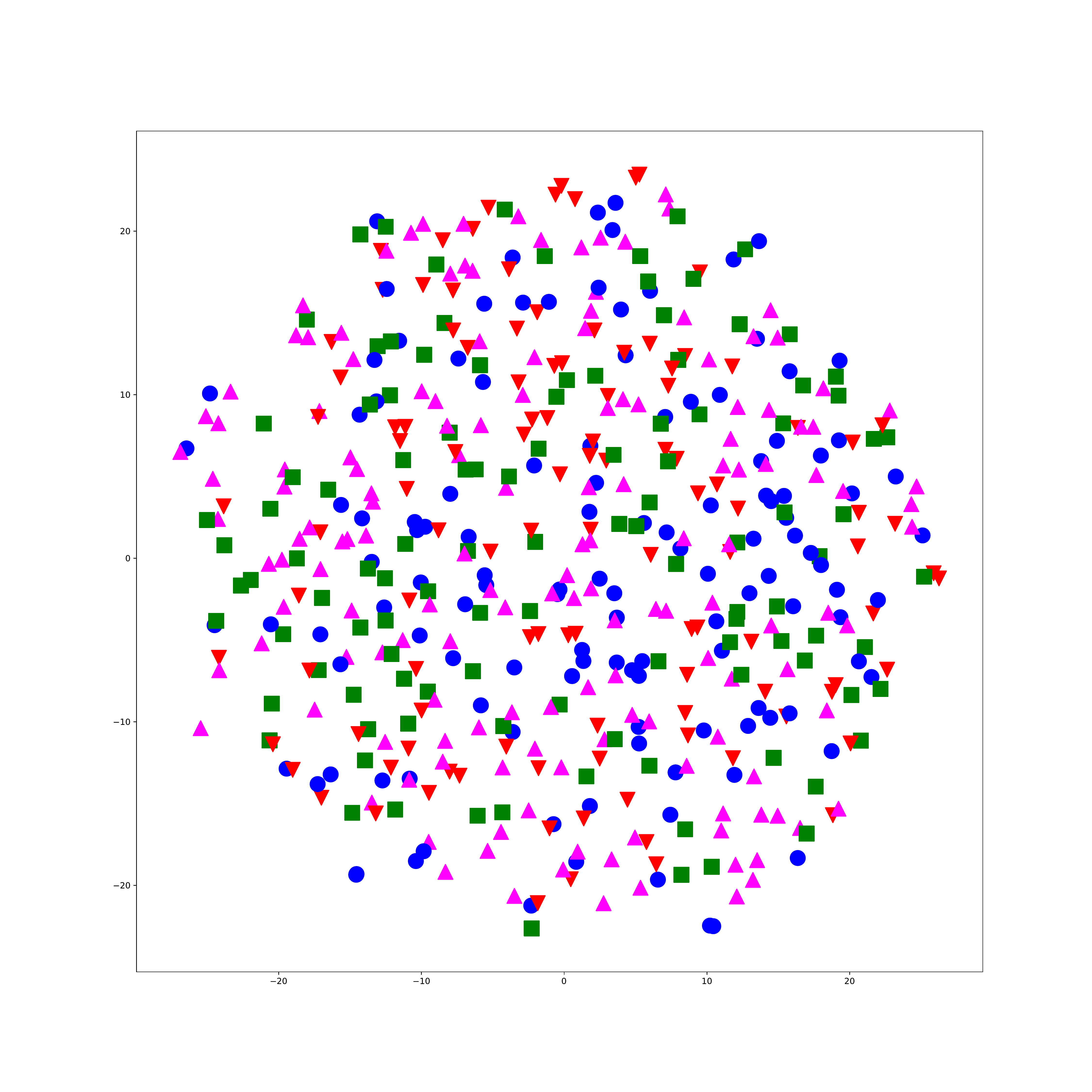}}\hspace{0em}%
  \subcaptionbox{Bootstrapping epoch 5\label{fig:bs_epoch_5}}{\includegraphics[width=0.3\textwidth]{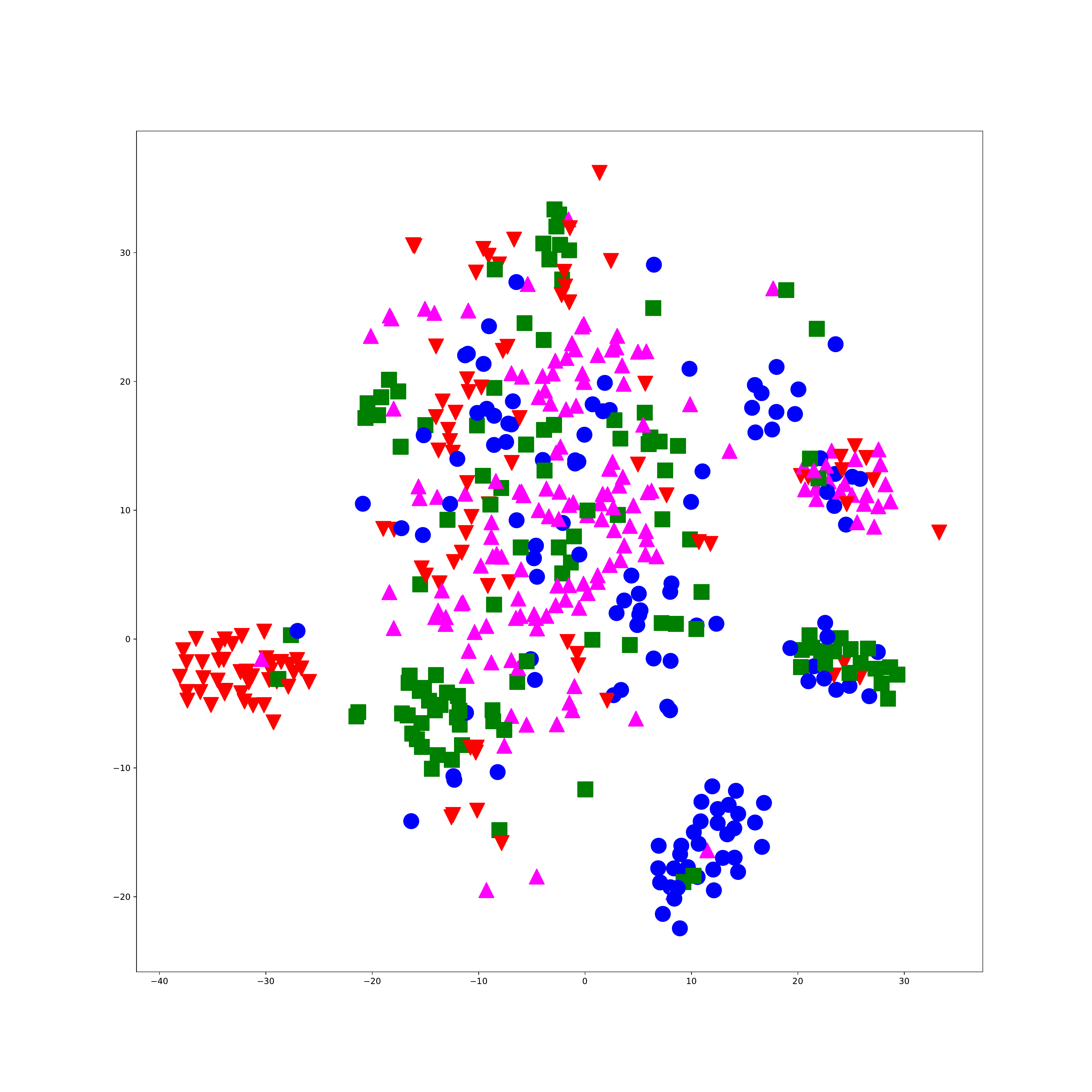}}\hspace{0em}%
  \subcaptionbox{Bootstrapping epoch 10\label{fig:bs_epoch_10}}{\includegraphics[width=0.3\textwidth]{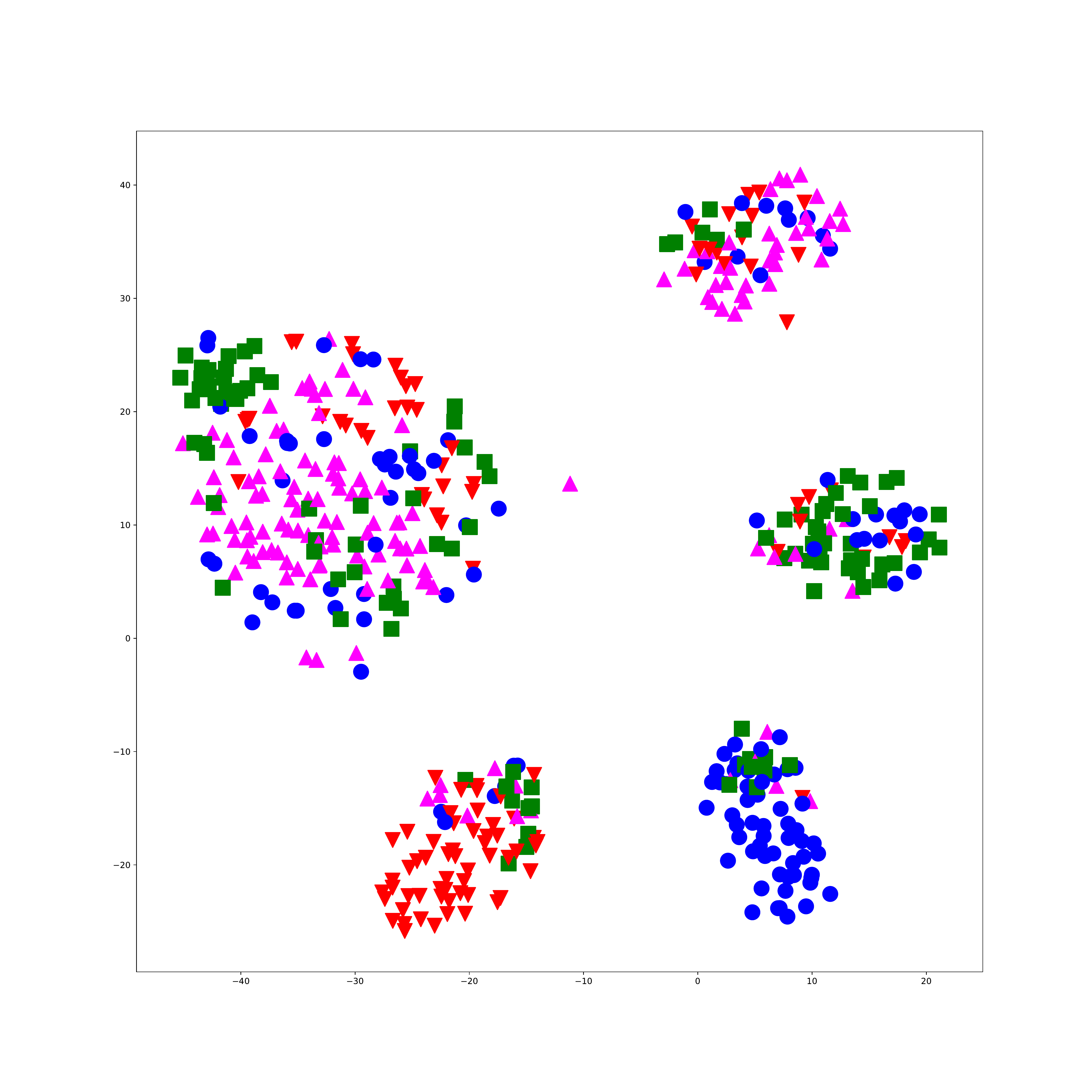}}\hspace{0em}%
  \vspace{-2mm}
    \caption{t-SNE visualizations of the entity embeddings at three stages during training. \\ Legend: \tikz\draw[blue,fill=blue] (0,0) circle (.5ex); = \texttt{LOC}. \tikz \fill [black!60!green](0.1,0.1) rectangle (0.3,0.3); = \texttt{ORG}. \tikz{\filldraw[draw=magenta,fill=magenta] (0,0) --
(0.2cm,0) -- (0.1cm,0.2cm);} = \texttt{PER}. \tikz{\filldraw[draw=red,fill=red, rotate=180] (0,0) --
(0.2cm,0) -- (0.1cm,0.2cm);} = \texttt{MISC}.}
  \label{fig:embeddings}
  \vspace{-4mm}
\end{figure*}

\begin{figure*}[th!]
\centering
\begin{tabular}{ c }  

\includegraphics[width=0.8\textwidth]{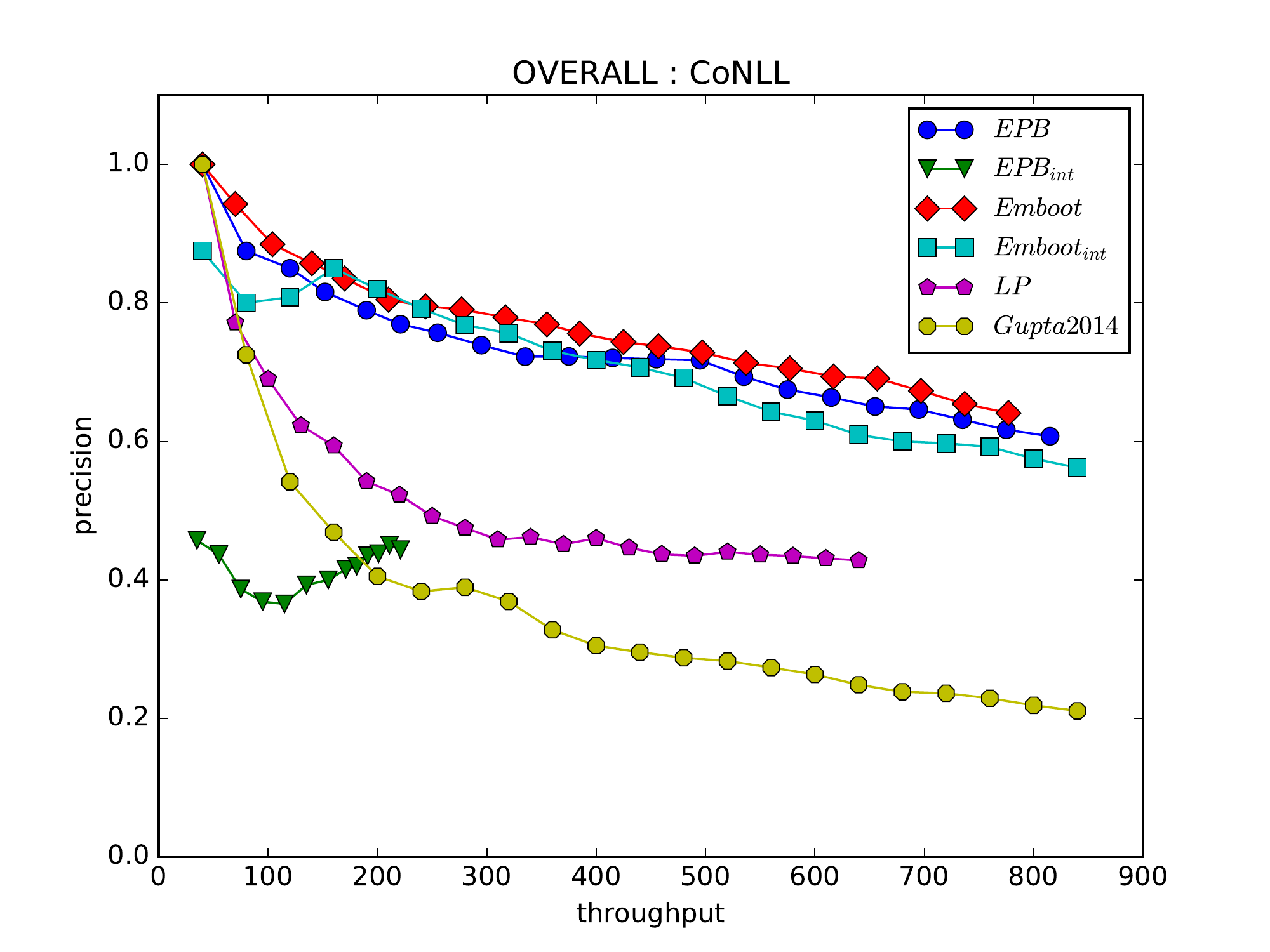}
\\
\includegraphics[width=0.8\textwidth]{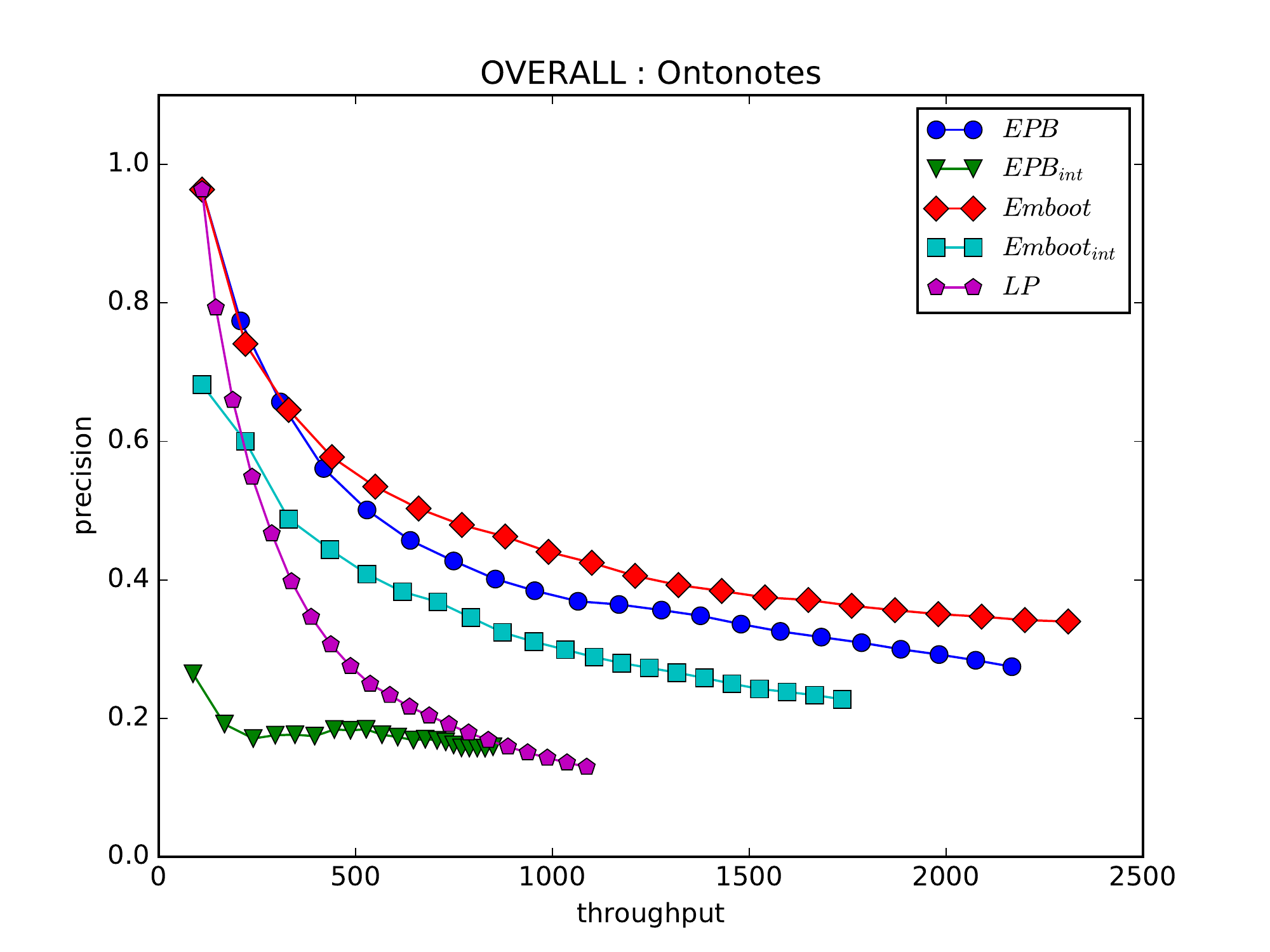} 

\end{tabular}  
\vspace*{-2mm}
\caption{\small Overall results on the  CoNLL and Ontonotes datasets. Throughput is the number of entities classified, and precision is the proportion of entities that were classified correctly. 
Please see Sec.~\ref{sec:experiments} for a description of the systems listed in the legend.
}
\label{fig:results}
\vspace{-2mm}
\end{figure*}

We evaluate the above algorithms on the task of named entity classification from free text.

{\flushleft {\bf Datasets}:} 
We used two datasets, the CoNLL-2003 shared task dataset~\cite{conll_dataset}, which contains 4 entity types, and the Ontonotes dataset~\cite{ontonotes}, which contains 11.\footnote{We excluded numerical categories such as \texttt{DATE}.}   
These datasets contain  marked entity boundaries 
with labels for each marked entity. Here we only use the entity boundaries but {\em not} the labels of these entities during the training of our bootstrapping systems. 
To simulate learning from large texts, we tuned hyper parameters on development, but ran the actual experiments on the {\em train} partitions. 

{\flushleft {\bf Baselines}:} In addition to the EPB algorithm, we compare against the approach proposed by~\citet{gupta2014improved}\footnote{\tiny{\url{https://nlp.stanford.edu/software/patternslearning.shtml}}}. 
This algorithm is a simpler version of the EPB system, where entities are promoted with a PMI-based formula rather than an entity classifier.\footnote{We did not run this system on Ontonotes dataset as it uses a builtin NE classifier with a predefined set of labels which did not match the Ontonotes labels.} 
Further, we compare against label propagation (LP)~\cite{zhu02learning}, with the implementation available in the {\tt scikit-learn} package.\footnote{\tiny{\url{http://scikit-learn.org/stable/modules/generated/sklearn.semi_supervised.LabelPropagation.html}}}
In each bootstrapping epoch, we run LP, select the entities with the lowest entropy, and add them to their top category. Each entity is represented by a feature vector that contains the co-occurrence counts of the entity and each of the patterns that matches it in text.\footnote{We experimented with other feature values, e.g., pattern PMI scores, but all performed worse than raw counts.}

{\flushleft {\bf Settings}:} For all baselines and proposed models, we used the same set of 10 seeds/category, which were manually chosen from the most frequent entities in the dataset. We used dependency-based word embeddings~\citep{depEmbed} of size 300 as the predefined embedding vectors in the entity promotion classifier. 
For the custom embedding features, we used randomly initialized 15d embeddings.
Here we consider patterns to be $n$-grams of size up to 4 tokens on either side of an entity. For instance, ``\texttt{@ENTITY , former President}'' is one of the patterns learned for the class {\tt person}. 
We ran all algorithms for 20 bootstrapping epochs, and the embedding learning component for 100 epochs in each bootstrapping epoch. 
We add 10 entities and 10 patterns to each category during every bootstrapping epoch.

\section{Discussion and Conclusion}

Before we discuss overall results, we provide a qualitative analysis of the learning process for Emboot for the CoNLL dataset in Figure~\ref{fig:embeddings}.
The figure shows t-SNE visualizations~\cite{van2008visualizing} 
of the entity embeddings at several stages of the algorithm. This visualization matches our intuition: as training advances, entities 
belonging to the same category are indeed grouped together.
In particular, Figure~\ref{fig:bs_epoch_10} shows five clusters, four of which are dominated by one category (and centered around the corresponding seeds), and one, in the upper left corner, with the entities that haven't yet been added to any of the pools.

A quantitative comparison of the different models on the two datasets is shown in Figure~\ref{fig:results}.  

Figure~\ref{fig:results} shows that Emboot considerably outperforms  LP and \citet{gupta2014improved}, and has a small but consistent improvement over EPB. This demonstrates the value of our approach, and the importance of custom embeddings. 

Importantly, we compare Emboot against: (a) its interpretable version (Emboot$_{int}$), which is constructed as a decision list containing the patterns learned (and scored) after each bootstrapping epoch, and (b) an interpretable system built similarly for EPB (EPB$_{int}$), using the pretrained Levy and Goldberg embeddings rather than our custom ones. 
This analysis shows that Emboot$_{int}$ performs close to Emboot on both datasets, demonstrating that most of the benefits of representation learning are available in an interpretable model. 
Further, the large gap between Emboot$_{int}$ and EPB$_{int}$ indicates that the custom embeddings are critical for the interpretable model.

Note that Emboot$_{int}$'s decisions are easy to interpret. Due to the sparsity of patterns, the majority of predictions are triggered by 1 or 2 patterns. For example, the entity ``Syrian'' is correctly classified as MISC (which includes  demonyms) due to two patterns matching it in the CoNLL dataset: ``{\tt @ENTITY President}'' and ``{\tt @ENTITY troops}''. In general, for the CoNLL dataset, 59\% of Emboot$_{int}$'s predictions are triggered by 1 or 2 patterns; 84\% are generated by 5 or fewer patterns; only 1.1\% of predictions are generated by 10 or more patterns. 

This work demonstrates that representation learning can be successfully combined with traditional, pattern-based bootstrapping, yielding models that perform well despite the limited supervision, and that are interpretable, i.e., end users can understand why an extraction was generated.

\bibliography{naacl2018}
\bibliographystyle{acl_natbib}

\end{document}